\documentclass[twocolumn,letterpaper]{IEEEAerospaceCLS}  

\usepackage[]{graphicx}    
\newcommand{\ignore}[1]{}  
\usepackage{color}
\usepackage{citesort}
\usepackage{url}
\usepackage{float}
\usepackage{epsfig} 
\usepackage{amsmath} 
\usepackage{amssymb}  
\usepackage{amsfonts}
\usepackage{url}

\DeclareMathAlphabet{\pazocal}{OMS}{zplm}{m}{n}
\newcommand{\Ws}{\pazocal{W}}
\newcommand{\Bs}{\pazocal{B}}
\newcommand{\Ts}{\pazocal{T}}
\newcommand{\Vs}{\pazocal{V}}

\begin{document}
\title{Visual--Thermal Landmarks and Inertial Fusion for Navigation in Degraded Visual Environments}

\author{%
Shehryar Khattak\\ 
Autonomous Robots Lab\\
University of Nevada, Reno\\
Reno, NV 89557\\
shehryar@nevada.unr.edu
\and
Christos Papachristos\\ 
Autonomous Robots Lab\\
University of Nevada, Reno\\
Reno, NV 89557\\
cpapachristos@unr.edu
\and 
Kostas Alexis\\
Autonomous Robots Lab\\
University of Nevada, Reno\\
Reno, NV 89557\\
kalexis@unr.edu
\thanks{\footnotesize 978-1-5386-6854-2/19/$\$31.00$ \copyright2019 IEEE}              
}

\maketitle

\thispagestyle{plain}
\pagestyle{plain}

\maketitle

\thispagestyle{plain}
\pagestyle{plain}

\begin{abstract}
During the past decade, aerial robots have seen an unprecedented expansion in their utility as they take on more tasks which had typically been reserved for humans. With an ever widening domain of aerial robotic applications, including many mission critical tasks such as disaster response operations, search and rescue missions and infrastructure inspections taking place in GPS--denied environments, the need for reliable autonomous operation of aerial robots has become crucial. To accomplish their tasks, aerial robots operating in GPS--denied areas rely on a multitude of sensors to localize and navigate. Visible spectrum camera systems correspond to the most commonly used sensing modality due to their low cost and weight rendering them suitable for small aerial robots in indoor or broadly GPS--denied settings. However, in environments that are visually--degraded such as in conditions of poor illumination, low texture, or presence of obscurants including fog, smoke and dust, the reliability of visible light cameras deteriorates significantly. Nevertheless, maintaining reliable robot navigation in such conditions is essential if the robot is to perform many of the critical applications listed above.
In contrast to visible light cameras, thermal cameras offer visibility in the infrared spectrum and can be used in a complementary manner with visible spectrum cameras for robot localization and navigation tasks, without paying the significant weight and power penalty typically associated with carrying other sensors such as 3D LiDARs or a RADAR. Exploiting this fact, in this work we present a multi--sensor fusion algorithm for reliable odometry estimation in GPS--denied and degraded visual environments. The proposed method utilizes information from both the visible and thermal spectra for landmark selection and prioritizes feature extraction from informative image regions based on a metric over spatial entropy. Furthermore, inertial sensing cues are integrated to improve the robustness of the odometry estimation process. The proposed method works in real-time, fully on-board an aerial robot. To verify our solution, a set of challenging experiments were conducted inside a) an obscurant-filed machine shop--like industrial environment, as well as b) a dark subterranean mine in the presence of heavy airborne dust.

\end{abstract}

\tableofcontents

\section{Introduction}\label{sec:intro}
Over the past decade aerial robots have seen an unprecedented expansion in the areas of their application as they provide an agile and flexible platform to perform human-reserved tasks autonomously, while minimizing the risk to human life and lowering operational costs. As a natural progression in this expansive role of aerial robots, they are now utilized in applications such as infrastructure inspection~\cite{roboticinspectionsurvey,ROSChapter,bircher2018receding,ICUAS2017,RHEM_ICRA_2017,VSEP_ICRA_2018,NasaUavElectrical2017,changeDetetion,icpMapping,m3u} and search and rescue missions~\cite{searchandrescure,tomic2012toward,thermalMarker,doherty2007uav,visionDepth}. These applications require robots to navigate through a variety of environments, while often encountering GPS--denied and visually degraded conditions making reliable pose estimation for the robot a major challenge. In the absence of external pose estimation such as GPS, which is common for indoor search and rescue operations, as well as inspection tasks of tunnel and mine environments, aerial robots rely on their on-board sensors to estimate odometry as they navigate an environment and perform their assigned tasks. For small unmanned aerial robots, visual spectrum camera sensors remain a popular choice as they provide a light-weight, power-efficient and low-cost sensor to enable odometry estimation as compared to 3D LiDAR and RADAR sensors. However, in environments with poor--illumination and low texture, image data from visual cameras become degraded and cannot be reliably used for pose estimation. Furthermore, the visual image quality suffers in the presence of visual obscurants such as smoke, fog and dust. In contrast to the limitations of visible spectrum cameras, thermal cameras operating in the Long Wave Infrared (LWIR) spectrum are not affected by conditions of poor-illumination and the presence of most visual obscurants, while maintaining the desired attributes of being light-weight and power-efficient. However, in general thermal images offer lower contrast in comparison to visual images especially when imaging areas containing objects of similar temperature and emissivity values and therefore require techniques such as histogram equalization to improve contrast~\cite{flir}. 

\begin{figure}[h!]
\centering
  \includegraphics[width=0.99\columnwidth]{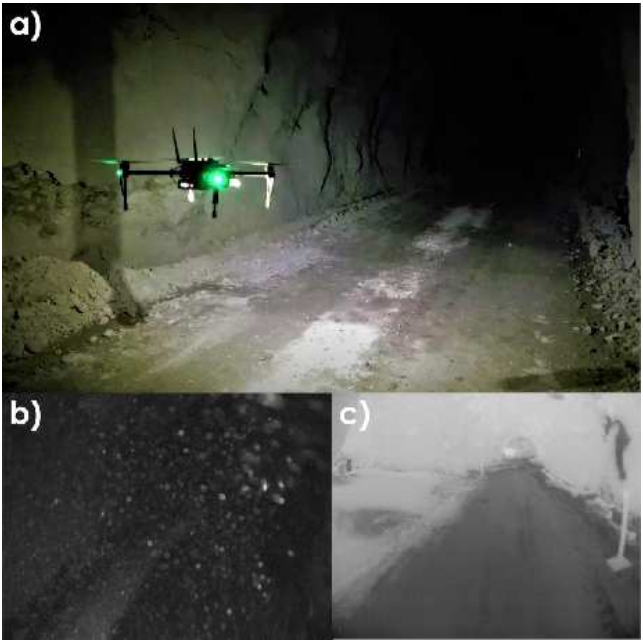}
\caption{Instance of an experiment during which an aerial robot navigates through a subterranean mine. Inset a) shows the aerial robot carrying the sensors and using on-board illumination, b) shows the visual image from the on-board visible camera, with low-illumination and the presence of airborne dust degrading the quality of the frame, while c) shows an image from the on-board thermal camera indicating that it remains unaffected by darkness and obscurants.}
\label{fig:mainfigure}
\end{figure}

Based on the discussion above, for aerial robots to reliably estimate their pose in a variety of challenging environments a multi--spectral odometry estimation approach, utilizing visual and thermal information in a complementary manner, can offer robustness when one sensing modality becomes degraded. In addition, such an approach is well suited for applications where aerial robots have to move between areas with different cases of sensor degradation, e.g. during search and rescue missions aerial robots might need to move from well--lit but thermally flat areas to smoke--filled areas and vice versa. Similarly, some obscurants might be temporarily introduced due to the operation of the aerial robot itself over different terrains, e.g. during underground mine inspection tasks, an aerial robot creates airborne dust due to the introduced air turbulence in often dry and dusty environments.
\par
Motivated by the aforementioned challenges, in this paper we present an approach to use entropy information in order to selectively utilize visual and thermal image information for the selection of landmarks for the pose estimation of aerial robots. These landmarks are then further fused with inertial measurements provided from an Inertial Measurement Unit (IMU) using an Extended Kalman Filter (EKF) framework. To demonstrate the real-time and on-board performance of our approach in degraded visual environments a set of experiments are conducted using an aerial robot, first in a fog-filled indoor environment. We demonstrate that the aerial robot can maintain a reliable estimate of its pose in the presence of fog, as well as while transitioning from well-lit to poorly-illuminated parts of the environment. Furthermore, we demonstrate the real world performance of our algorithm by estimating the pose of the aerial robot inside a subterranean mine environment in the absence of external illumination and in the presence of heavy airborne dust. A video of the conducted experiments can be found at \url{https://youtu.be/aqZugneeCxc}.
Figure~\ref{fig:mainfigure} presents an instance of an underground mine exploration mission. 

The remainder of the paper is structured as follows: Section~\ref{sec:related} provides an overview of the related work, while Section~\ref{sec:approach} details our proposed method. The experimental evaluation studies are presented in Section~\ref{sec:experiments}. Finally, conclusions are drawn in Section~\ref{sec:conclusion}.

\section{Related Work}\label{sec:related}
Estimation of vehicle pose in Degraded Visual Environments (DVE) has been a widely studied topic in the context of navigation and landing of helicopters especially in conditions of darkness and in the presence of heavy airborne dust. Solutions proposed for larger aerial vehicles are usually based on LiDAR~\cite{heliLIDARObscurant} or RADAR~\cite{heliRADARObscurant} systems given the higher payload capacity of such vehicles and how cost factors are considered. However, given the smaller payload capacity of Micro Aerial Vehicles (MAVs), LiDAR and especially RADAR system integration comes with high power and operational time penalties. Considering the case of lightweight sensing solutions carried on-board aerial robots, most solutions emphasize on visual--inertial localization and only more recently the community has considered the potential of thermal vision. The work in~\cite{vidas} showed a hand--held setup consisting of an RGBD and a thermal camera for thermographic mapping of objects. However, the underlying odometry estimation relies purely on RGBD data. Similarly, in~\cite{chameleon} a hand--held system consisting of visible and thermal cameras along with an IMU was presented. Authors demonstrated the feasibility of thermal and IMU only odometry estimation, however they also showed the limitations of such a system in areas with very low thermal gradients, while also showing visual odometry to work in the same environment and concluded by suggesting the usage of a multi--spectral approach for odometry estimation. In~\cite{practicalTO2016} authors demonstrate odometry estimation of a car using a monocular thermal camera by extending visual odometry approaches but discuss the necessity of knowing the height of camera from the ground at all times to estimate the scale correctly. The authors also discuss the lack of contrast in thermal images and suggest improving the contrast in thermal images, as well as emphasizing the need for the fusion of thermal imagery with other sensing modalities to make odometry estimation more robust. A stereo thermal camera system is proposed in~\cite{thermalstereo2015} for the estimation of odometry with correct scale. The authors demonstrate the pose estimation of an aerial robot with this system in outdoor settings but discuss the need of integration of proprioceptive sensors such as an IMU to make the overall odometry solution robust.

Following the direction of a multi--spectral odometry estimation solution that utilizes both visible and thermal information, the authors in~\cite{rgbt} propose detecting features in both visual and thermal images for odometry estimation. They demonstrate the feasibility of their approach by estimating odometry under different lighting conditions in outdoor settings. Their method proposes separate detection of features in visual and thermal images but extraction of feature descriptors in both domains for each feature point for robust feature matching. However, as discussed in~\cite{thermalFeaturePerformance} the direct application of feature detection and descriptor matching methods designed for visual images on thermal camera data results in poor feature matching performance and reduced accuracy of the estimated odometry. Echoing the same argument, the authors in~\cite{mutlispectral2015} propose an edge--histogram based descriptor for multi--spectral feature matching on a visual and thermal camera stereo system. They demonstrate the application of their method by estimating the odometry of a car navigating through an urban environment. However, as discussed in the experimental section of the paper, the proposed descriptor shows inferior feature matching performance when applied to visual images and is replaced with typical visual methods during experiments. Similarly, their odometry estimation approach integrates pose priors provided by GPS making their solution unsuitable for GPS--denied environments. In~\cite{selectiveVisualThermal2013}, authors estimate the pose of an unmanned ground vehicle using visual and thermal cameras in the presence of a variety of obscurants in outdoors settings. Features are selectively chosen from areas of visual and thermal images which are not degraded by the presence of obscurants and are used to estimate the robot pose. Scale is calculated independently by using velocity information provided by the IMU. The authors also discuss poor feature matching performance on thermal images, as well as in the presence of obscurants, and suggest the application of methods such as RANSAC, and its variations, to prune incorrect feature matches for improving the reliability of odometry estimation.

Motivated by progress in the literature and the need to address the identified limitations of the current state-of-the-art, in this paper we propose a multi--modal, multi--spectral approach on visual/thermal camera and IMU fusion. The proposed solution utilizes spatial entropy to prioritize certain visual and thermal image regions for feature selection and fuses IMU cues in an EKF fashion. We demonstrate the strengths of our solution in GPS-denied, dark and obscurants-filled environments.

\section{Proposed Method}\label{sec:approach}
\begin{figure}[h!]
\centering
  \includegraphics[width=0.99\columnwidth]{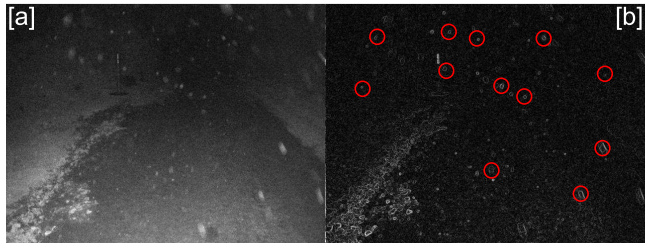}
\caption{Effect of obscurants on sensor degradation for robot localization: [a] and [b] show the visible image and its calculated gradient image respectively. Dust particles seen in [a], create corresponding strong gradient values in [b] as highlighted by the red circles. These strong gradients are temporal in nature and lead to selection of features that cannot be tracked in successive frames.}
\label{fig:dustgradients}
\end{figure}

To enable reliable odometry estimation for aerial robots in degraded visual environments (DVEs) using visible and thermal cameras, the selection of interest points (or features) must be made carefully. As discussed in Section~\ref{sec:related}, the application of visual feature detection and matching methods on images in the presence of obscurants can lead to poor estimation of odometry because of feature mismatching. This is due to the fact that feature detection and matching methods operate over different neighborhood sizes in an image with the detection of features being performed first by evaluating the gradient information over a small neighborhood at each pixel location in the image. Once a set of features has been selected, a feature descriptor in feature--based methods, or an intensity patch in direct intensity methods, is then extracted to encode information over a larger neighborhood around each feature location for matching and tracking in subsequent frames. As the neighborhood sizes for evaluation are different in the detection and matching operations, a feature may be selected which is associated with a non-informative larger neighborhood for matching operations. Such an incorrect selection of features can easily be made in the presence of obscurants such as fog and smoke, which unevenly decrease the quality of gradient information in an image~\cite{selectiveVisualThermal2013}. 
Furthermore, obscurants such as airborne dust can also temporarily introduce strong local gradients which can lead to the selection of non-repeatable features.
This may lead to tracking failure or incorrect estimation of confidence metrics for an odometry solution. An example of temporary strong local gradients introduced by dust particles is shown in Figure~\ref{fig:dustgradients}, where dust speckles introduce strong local gradients against a dark background and can be incorrectly selected as features.  

\begin{figure}[h!]
\centering
  \includegraphics[width=0.99\columnwidth]{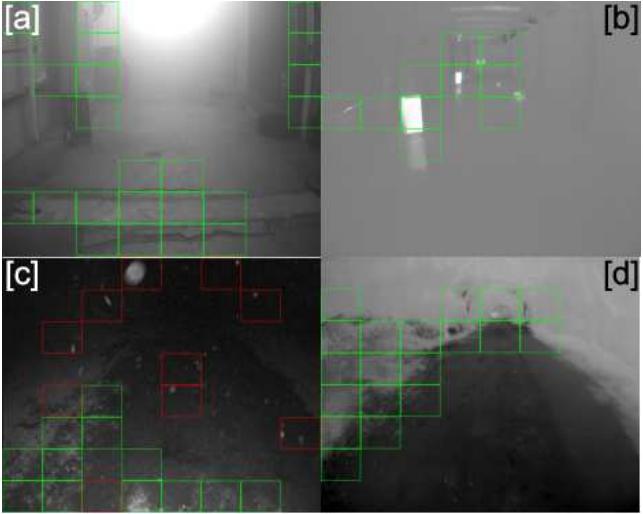}
\caption{Figure shows the proposed feature selection strategy applied to visible and thermal images collected in indoor and subterranean mine environments. Inset [a] and [b] show the visible and thermal images respectively, collected in an indoor environment in the presence of dense fog. Green boxes show the informative regions of the image from where features should be selected. Similarly, [c] and [d] show the visible and thermal images respectively, collected in a subterranean mine environment in the presence of heavy airborne dust. Red boxes show rejected regions containing dust due to their temporally inconsistent quality.}
\label{fig:indoor_mine_mask}
\end{figure}

In our approach we associate the local feature selection process to larger image neighborhoods to make sure that the selected features have corresponding informative neighborhoods in order to ensure better feature tracking performance. For this purpose we first measure the amount of information contained in an image which has traditionally been done by using the Shannon Entropy criterion, given as:

\begin{eqnarray}\label{entropyimage}
e_{image} = - \sum_{i}^{N}\vec{p}_{i}\cdot \log(\vec{p}_{i})
\end{eqnarray}
where $e_{image}$ is the entropy measure of the image and $\vec{p}$ is the probability distribution vector of image intensity values over all image pixels $N$ in the image. However, we are interested in measuring the amount of spatial structure present in an image as features typically lie at edge or corner points of objects in the frame. Hence, instead of using direct image intensity values for entropy calculation we make use of the edge-filtered gradient image as it better encodes the underlying spatial structure present in an image as shown in~\cite{ThermalPerception}. Moreover, instead of calculating a global metric for an image we divide the image into $R \times R$ regions and calculate the spatial entropy of each region. This division allows us to better preserve local edges in the gradient image which otherwise might have been lost when subjected to a global threshold. Furthermore, although spatial entropy can provide a measure of the structural information available in an image region, it still remains context--free, as a region with an equal amount of very small and very large gradients produces the same spatial entropy measure as a region with a uniform distribution of gradient values as detailed in~\cite{imageentropy}. Since we are interested in selecting features from regions of the image with a more uniform distribution of gradients, we further introduce a weight vector to weigh our entropy vector using a Gaussian distribution. Our entropy equation for an image region can now be written as:

\vspace{-5ex}
\begin{eqnarray}\label{entropyregion}
e_{region} = -\sum_{i}^{N_R}(\vec{\omega}\odot\vec{p}_{i})\cdot \log(\vec{p}_{i})
\end{eqnarray}
where $e_{region}$ is the entropy measure of the region, $\vec{\omega}$ is the Gaussian weight vector, $\odot$ represents the Hadamard product between two vectors and $\vec{p}_{i}$ represents the probability distribution vector of gradient magnitude values over all pixels $N_{\mathbf{R}}$ in a certain image region $\mathbf{R}$. Once the region--wise spatial entropy of the image has been calculated we compute the mean spatial entropy among all the regions and mark regions having an entropy measure above the mean as candidate regions from which features can be selected. Furthermore, in order to account for the dynamic nature of obscurants, we track the spatial entropy of a region over a temporal window and allow the selection of features from a region only if the region remains informative over a select number of frames. Figure~\ref{fig:indoor_mine_mask} shows the application of the combined spatial and temporal quality metric on visible and thermal images, where green boxes indicate candidate regions for feature selection, red boxes show rejected regions which were labeled as candidate regions in the current frame but rejected due temporally inconsistent quality.

\begin{figure}[h!]
\centering
  \includegraphics[width=0.99\columnwidth]{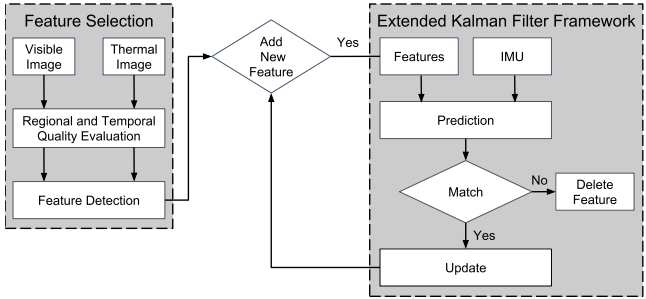}
\caption{Overview of the proposed approach.}
\label{fig:mask_rovio}
\end{figure}

Once features have been selected from the informative regions of visible and thermal images, they are fused with inertial measurements using an Extended Kalman Filter (EKF) for the odometry estimation of the aerial robot. A schematic diagram of our approach is shown in Figure~\ref{fig:mask_rovio}.

We use an EKF formulation similar to the one proposed in~\cite{rovio} and track features as part of our filter state by modeling them as $3\textrm{D}$ landmarks using a $2\textrm{D}$ bearing vector $\boldsymbol{\mu}$, parameterized with azimuth and elevation angles, and an inverse depth parameter $\boldsymbol{\rho}$. The filter state propagation is performed using proper acceleration $\hat{\mathbf{f}}$ and rotational rate measurements $\hat{\boldsymbol{\omega}}$ provided by the IMU. Propagating the filter state with IMU readings allows the method to predict the feature locations between successive frames, reducing the search space for feature matching in the filter update step and negates the need for a separate feature mismatch pruning step. Feature matching is performed by minimizing the photometric error of a neighborhood patch at each feature's predicted location. In our formulation, four coordinate frames namely, the IMU fixed coordinate frame ($\Bs$), the visible camera fixed frame $\Vs$, the thermal camera fixed frame $\Ts$, and the world inertial frame $\Ws$, are used. It should be noted that we do not separately reserve features in the state vector for either visible or thermal images and choose the best available feature across spectra when selection of a new feature is needed. Permitting interchangeable features to be tracked as part of the filter state allows the method to keep the filter state vector small and therefore remain computationally efficient. The employed state vector of our EKF formulation takes the form:

\vspace{-4ex}
\begin{eqnarray}\label{eq:roviostate}
 \mathbf{x} = [ \underbrace{\mathbf{r}~\mathbf{q}~\boldsymbol{\upsilon}~\mathbf{b}_f~\mathbf{b}_\omega
 }_\text{robot states}~|~\underbrace{\boldsymbol{\mu}_0,~\cdots~\boldsymbol{\mu}_J~\rho_0~\cdots~\rho_J}_\text{landmark states}]^{\top}
\end{eqnarray}
where $\mathbf{r}$ and $\boldsymbol{\upsilon}$ are the position and velocity of the IMU respectively, expressed in the $\Bs$ coordinate frame, $\mathbf{q}$ is the IMU attitude from the $\Bs$ to the $\Ws$ coordinate frames, $\mathbf{b}_f$ and $\mathbf{b}_\omega$ represent the additive accelerometer and gyroscope biases respectively expressed in the $\Bs$ coordinate frame,
while $\boldsymbol{\mu} _j$ is the bearing vector to the feature $j$ expressed in the $\Vs$ or the $\Ts$ coordinate frame and $\rho_j$ is the inverse depth parameter of the $j^{th}$ feature such that the feature distance $d_j$ is $d(\rho_j) = 1/\rho_j$. 

\section{Experimental Evaluation}\label{sec:experiments}
To evaluate the performance of our proposed algorithm, a set of experimental studies were conducted using an aerial robot in a) an indoor machine shop environment subject to the presence of dense fog and partial darkness, as well as b) a subterranean mine environment in the presence of heavy airborne dust. The robotic system used and the experiments conducted are detailed below. A video of the relevant experiments can be found at \url{https://youtu.be/aqZugneeCxc}

\begin{figure}[h!]
\centering
  \includegraphics[width=0.99\columnwidth]{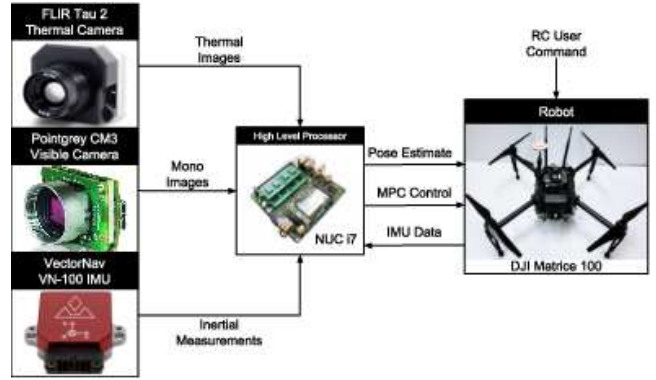}
\caption{The high-level processor, Intel NUC--i7, receives visual, thermal and inertial data from the Chameleon 3 camera, the Tau 2 thermal camera and the VN--100 IMU respectively in order to reliably estimate the robot pose. Furthermore, the high-level processor also receives inertial measurements from the autopilot of the robot and estimates smooth and robust robot odometry in real--time to enable state feedback position control.}
\label{fig:system}
\end{figure}

\begin{figure*}[h!]
\centering
  \includegraphics[width=\textwidth]{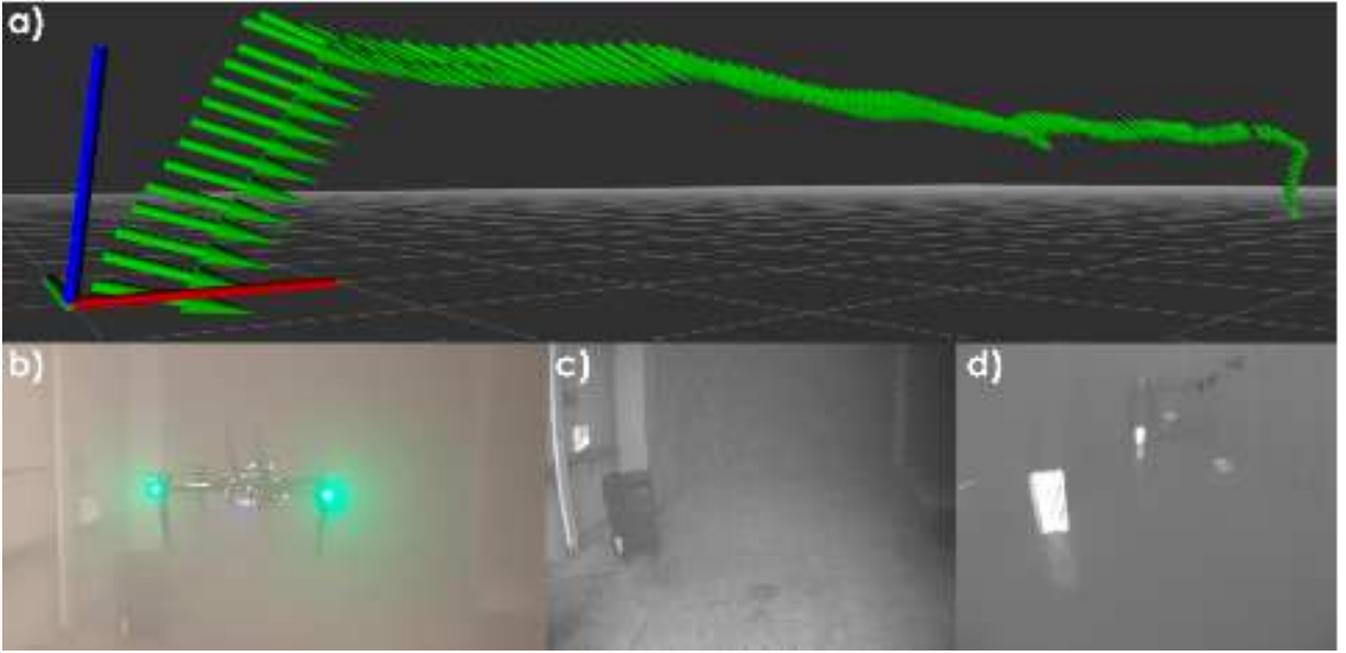}
\caption{A visualization of the robot trajectory during the indoor experiment. Inset a) shows the robot trajectory in green, b) shows an image of the robot during an instance of the experiment navigating from lit to dark parts of the environment, while c) and d) show images from the on-board visible and thermal cameras respectively at this instance during the robot trajectory.}
\label{fig:indoors_rviz}
\end{figure*}

\subsection{System Overview}
For the purposes of experimental evaluation, a DJI Matrice 100 quadrotor equipped with the sensing suite proposed in this paper and an on-board computer was used. A FLIR Tau 2 thermal camera was mounted on the robot to provide thermal images of $640\times512$ resolution at $30$ frames per second. Two Pointgrey Chameleon 3 visible light CCD cameras, with shutter--synchronized LEDs, provided visual images of $644\times482$ resolution at $20$ frames per second. Only one of the visual cameras alongside the thermal camera was used during the experimental evaluations of this work. A VectorNav VN--100 IMU was employed to provide inertial measurements at an update rate of $200\textrm{Hz}$. The intrinsic calibration parameters of the visual and thermal cameras, as well the extrinsic calibration parameters for the IMU and the cameras were estimated using the work in~\cite{kalibr} and our custom designed thermal checkerboard pattern~\cite{ICUAS2018Thermal}. An Intel NUC--i7 computer (NUC7i7BNH) was carried on-board the robot to perform all high-level processing tasks. The estimated odometry from our proposed method is fused with inertial measurements received from the low--level attitude controlling autopilot of the robot using the the work of~\cite{msf}, in order to enable smooth and reliable position control using a Linear Model Predictive Control strategy based on the previous work in~\cite{mpc_rosbookchapter}. All algorithms were implemented as Robot Operating System (ROS) nodes and run in real-time fully on-board the robot with the odometry estimation operation performing at an update rate of $20\textrm{Hz}$. Figure~\ref{fig:system} provides the system overview of the robotic system used.



\subsection{Indoor Experiments}
To evaluate the real--time on-board performance of the proposed approach on an aerial robot, flight tests were conducted in an obscurant-filled environment. A fog generator was used to fill a machine shop resembling industrial environment with fog to serve as an obscurant during the experiment. Furthermore, lights were turned off in a section of the environment to simulate partial darkness and to demonstrate the robustness of our approach under varying light conditions and in the presence of obscurants during the same experiment. It should be noted that no on-board illumination was used in these experiments. The estimated robot trajectory during flight, from take--off to landing, is shown in Figure~\ref{fig:indoors_rviz}.

\begin{figure}[h!]
\centering
  \includegraphics[width=0.99\columnwidth]{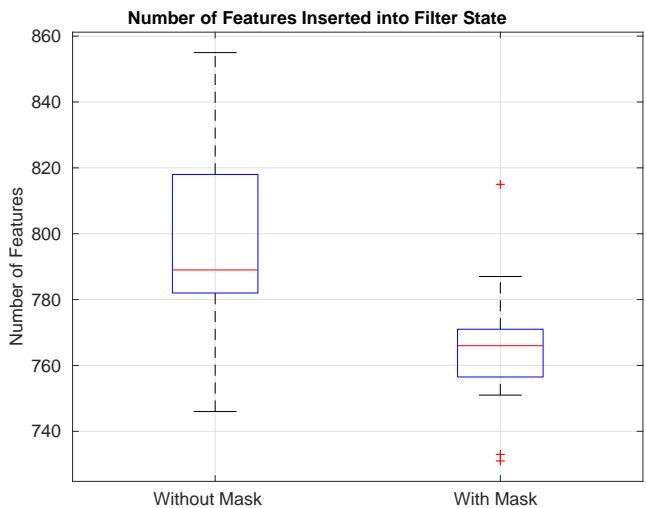}
\caption{Plot showing the total number of features inserted into the filter state, with and without the application of our proposed feature selection strategy, during the indoor experiments. Red line indicates the median value in each population, the blue box represents the interquartile range between the first and third quartile, while red $+$ symbols represent the outliers.}
\label{fig:indoor_boxplot}
\end{figure}

To understand the effect of our feature selection approach on the track--ability of selected features inserted into the filter sate, we conducted 10 trials with and without using our feature selection region mask on the collected data-sets during the indoor experiments. The results of these trials are shown in Figure~\ref{fig:indoor_boxplot}. The plot demonstrates the benefit of our feature selection approach as the median number of features inserted in the filter state is smaller, indicating that the features selected using our proposed approach are tracked for a greater number of frames. Furthermore, to verify that our feature selection strategy provides a statistically significant difference in the selection of features we performed one--way Analysis of Variance (ANOVA) test on the population of features selected with and without the application of our approach. The test resulted in $p\text{-}value < 0.007185$, indicating that the two populations are statistically different from each other and verifies the effect and impact of our proposed feature selection strategy.

\begin{figure}[htb]
\centering
  \includegraphics[width=0.99\columnwidth]{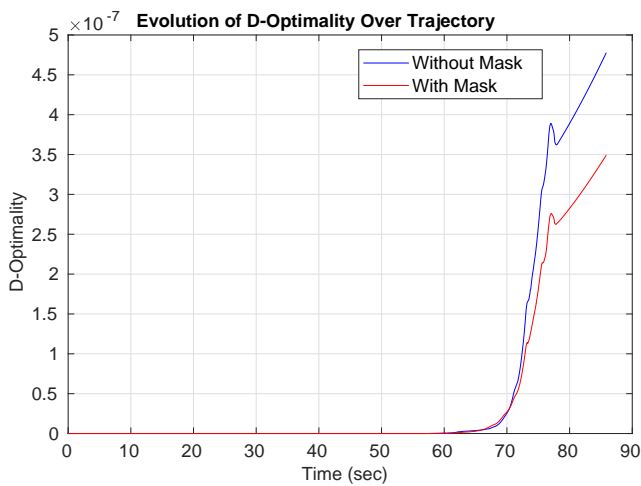}
\caption{Plot showing the evolution of the robot pose uncertainty, using the D-Optimality criterion, with and without the application of our proposed feature selection strategy during an indoor experiment.}
\label{fig:indoor_dopt}
\end{figure}

Likewise, to understand the effect of our proposed feature selection strategy on the quality of the robot pose estimates, we analyze the evolution of the associated covariance of the robot pose estimates over time. As recommended in~\cite{dopt}, we employ the D-Optimality criterion as a metric for measuring uncertainty of our robot pose estimates, with a lower D-Optimality value indicating a lower uncertainty in the robot pose estimate. This takes the form:

\vspace{-2ex}
\begin{eqnarray}\label{dopt}
D_{Optimality} = \exp(\log(\det(\Sigma)^{1/l}))
\end{eqnarray}
where $\Sigma$ is the robot pose covariance matrix with dimensions $l\times l$. The evolution of the D-Optimality criterion for the robot pose estimates over the length of the mission is shown in Figure~\ref{fig:indoor_dopt}. It can be noted that the application of our feature selection mask attenuates the accumulation of uncertainty in robot pose estimates, as features that can be tracked for more successive frames lead to better convergence of filter uncertainty estimates.

\subsection{Subterranean Mine Experiment}
\begin{figure*}[h!]
\centering
  \includegraphics[width=\textwidth]{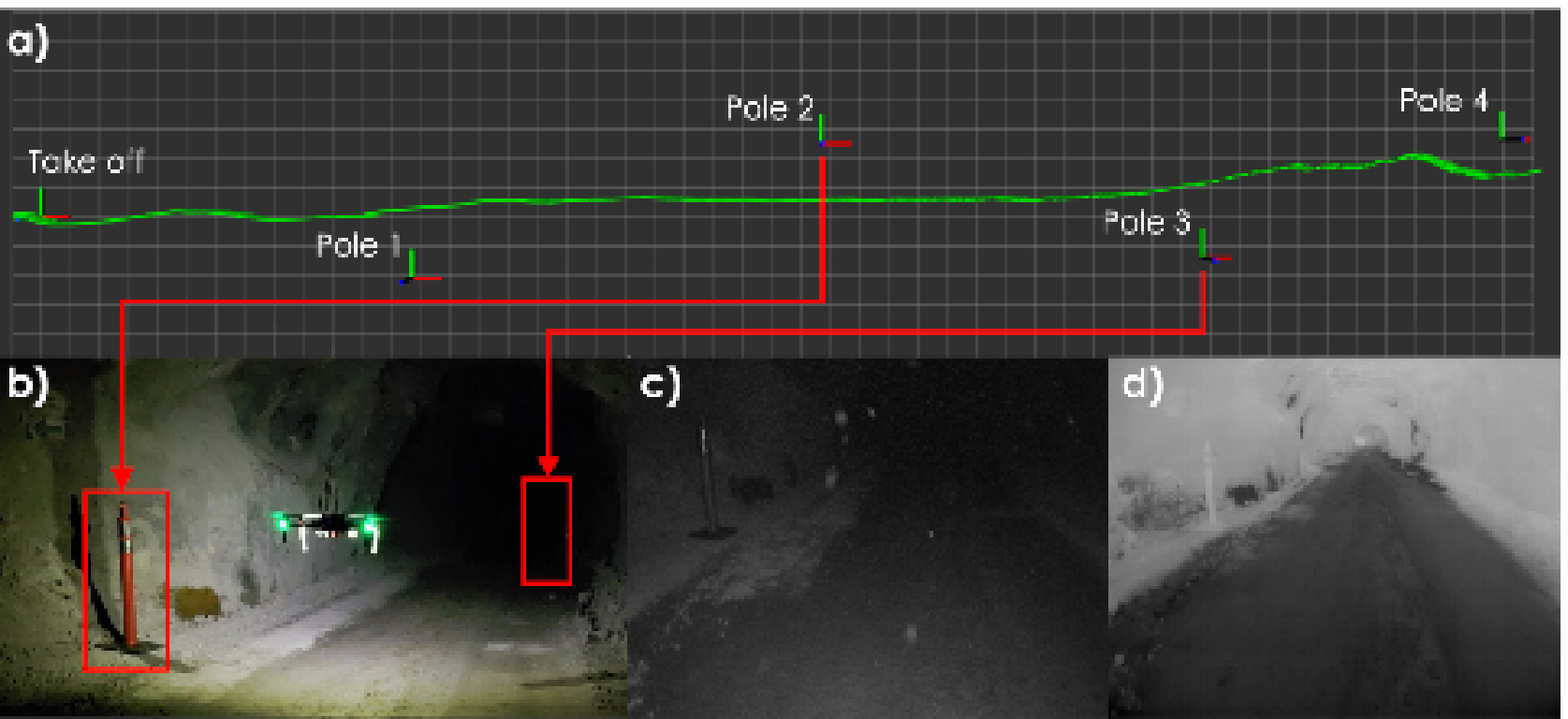}
\caption{A top--down view of the robot trajectory as well as the position of pole markers placed to provide a measure of ground--truth during the subterranean mine experiment are shown in a), an image of the robot and the two observed pole markers during an instance of this experiment are shown in b), while c) and d) show the images captured from the on--board visual and thermal cameras respectively during an instance of the subterranean mine experiment.}
\label{fig:mine_rviz}
\end{figure*}

To demonstrate the real world performance and application of our approach, an experiment was conducted in a subterranean mine environment in conditions of darkness and heavy airborne dust. On-board illumination was utilized for visible light cameras to function in the otherwise completely dark underground mine environment. To provide a measure for ground--truth, pole markers were placed along the direction of the mine shaft with the final marker position being $50.0\textrm{m}$ away from the robot take-off position. The estimated robot trajectory during the experiment is shown in Figure~\ref{fig:mine_rviz}. A final position error of $0.4857\textrm{m}$ was estimated for the robot navigation along the direction of the mine shaft. 

\begin{figure}[h!]
\centering
  \includegraphics[width=0.99\columnwidth]{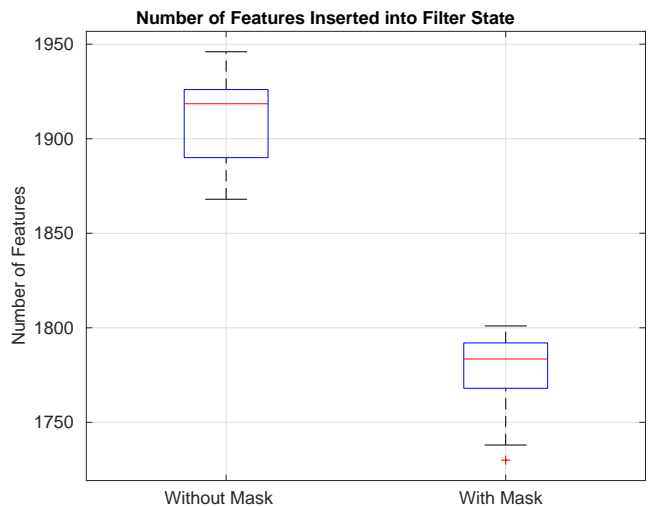}
\caption{Plot showing the total number of features inserted into the filter state, with and without the application of our proposed feature selection strategy, during the subterranean mine experiments. Red line indicates the median value in each population, the blue box represents the interquartile range between the first and third quartile and red $+$ symbols represent the outliers.}
\label{fig:mine_boxplot}
\end{figure}

Similar to the analysis of indoor experiments, we conducted 10 trials on the collected data during the experiment to verify the effect of our approach. The respective plot is shown in Figure~\ref{fig:mine_boxplot}. As shown, the median number of features inserted into the filter state during the mission length using the proposed feature selection mask is lower as compared to not using it. Similarly, an ANOVA test was again performed on the two populations resulting in a $p\text{-}value < 0.00001$, validating that our approach makes a statistically significant difference on the selection of features to be used for estimation of odometry for the aerial robot.

\begin{figure}[h!]
\centering
  \includegraphics[width=0.99\columnwidth]{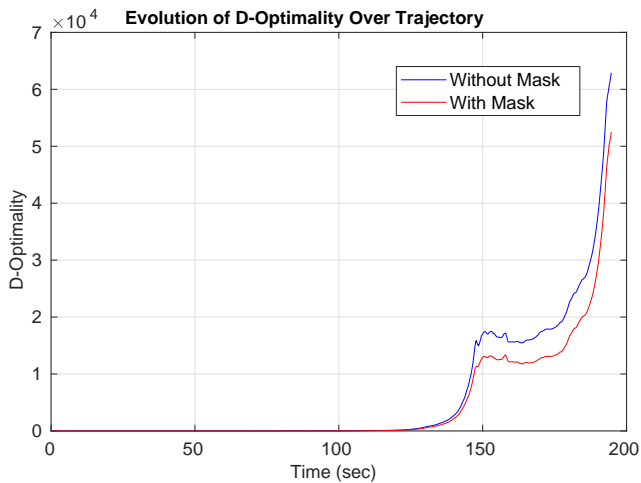}
\caption{Plot showing the evolution of the robot pose uncertainty, using the D-Optimality criterion, with and without the application of our proposed feature selection strategy during a subterranean mine experiment.}
\label{fig:mine_dopt}
\end{figure}

Similarly, Figure~\ref{fig:mine_dopt} shows the evolution of the pose uncertainty, using the D-Optimality criterion, over the trajectory of the aerial robot navigating through the subterranean mine environment. It is highlighted that the application of our feature selection mask results in a slowdown of the growth of pose uncertainty.

\section{Conclusions}\label{sec:conclusion}
In this paper, a solution to enable autonomous navigation of aerial robots in environments that are not only GPS--denied but also visually--degraded is presented. A multi--modal sensor fusion approach is proposed and relies on the fusion of visible light and thermal vision sensors, alongside inertial measurement cues. The algorithm utilizes information from both the visible and thermal spectra for landmark detection and prioritizes feature extraction from informative image regions based on a metric over spatial entropy. Through extensive experimental evaluation we demonstrate that the described solution enables reliable and accurate robot localization in challenging GPS--denied, possibly subterranean, dark, textureless and obscurants--filled environments including underground mines. 

\acknowledgments
This material is based upon work related to the Mine Inspection Robotics project sponsored by the Nevada Knowledge Fund administered by the Governor's Office of Economic Development.

\bibliographystyle{IEEEtran}
\bibliography{AEROCONF2019}

\thebiography
\begin{biographywithpic}
{Shehryar Khattak}{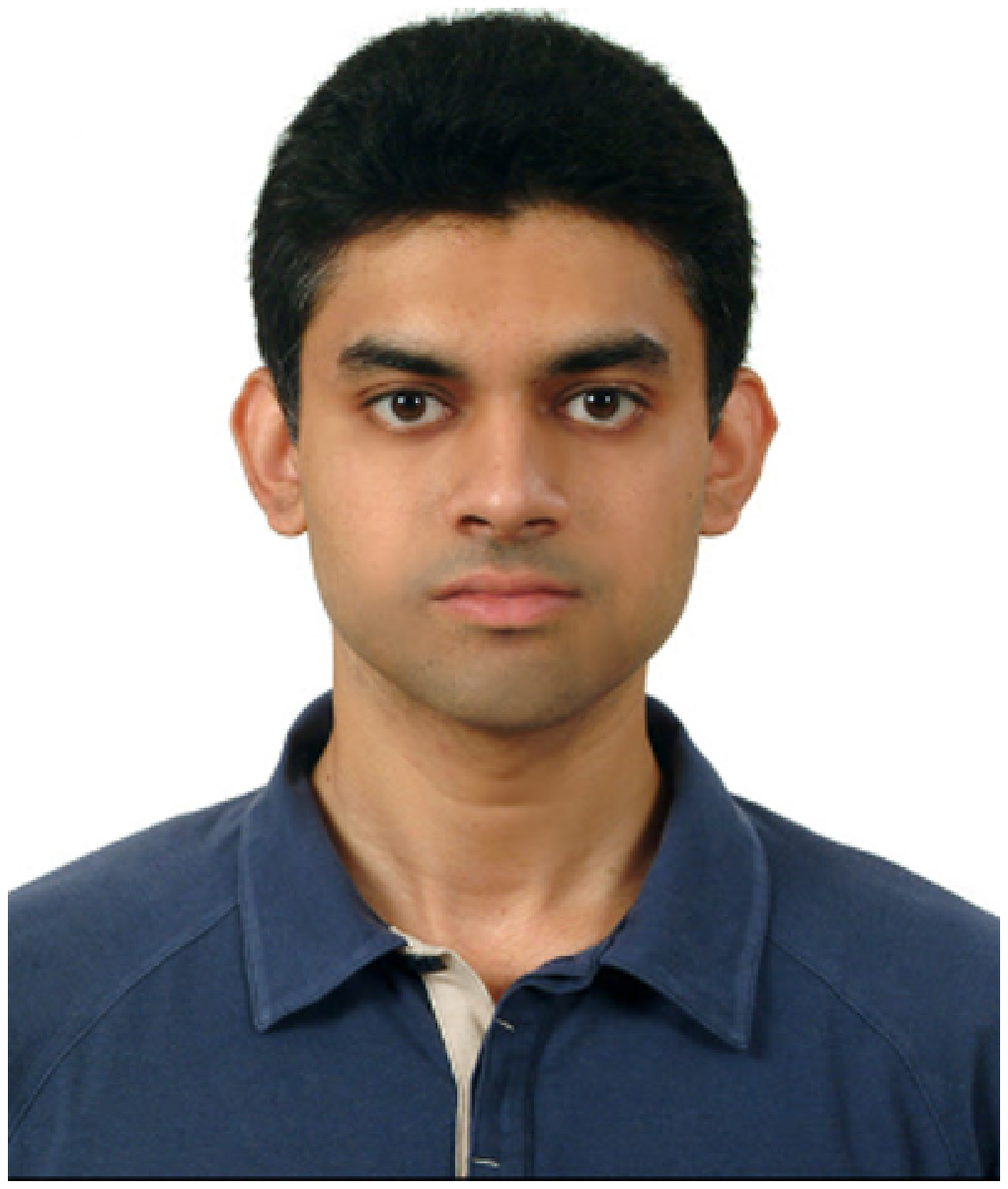} 
earned his B.S. in Mechanical Engineering from Ghulam Ishaq Khan Institute of Engineering Sciences and Technology, Pakistan in 2009 and M.S. in Aerospace Engineering from Korea Advanced Institute of Science and Technology, Daejeon in 2012. From August 2012 to December 2015, he worked as a Research Engineer at Samsung Electronics in Suwon, South Korea. Currently, Shehryar is pursuing his Ph.D. in Computer Science and Engineering from the University of Nevada, Reno. His current research is related to robot perception and path planning with focus on development of localization and mapping algorithms exploiting multi-sensor information.
\end{biographywithpic}
\begin{biographywithpic}
{Christos Papachristos}{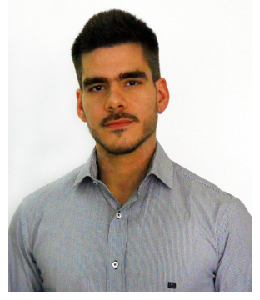} 
is a Post-Doctoral Researcher at the Autonomous Robots Lab of the University of Nevada Reno. He works in the fields of aerial robotics design, control, perception and planning, focusing on the systems and the methods that promote autonomy, long-term reliability, and robustness of operation in unknown environments. His current research activities include enabling GPS-denied localization for Micro Aerial Vehicles operating in complex degraded visual environments, and achieving their autonomous exploration, mapping, and characterization. Christos Papachristos received his PhD from the University of Patras, Greece in 2015, and has since worked in large-scale research projects that aim to advance robotic autonomy, on both sides of the Atlantic. 
\end{biographywithpic}
\begin{biographywithpic}
{Kostas Alexis}{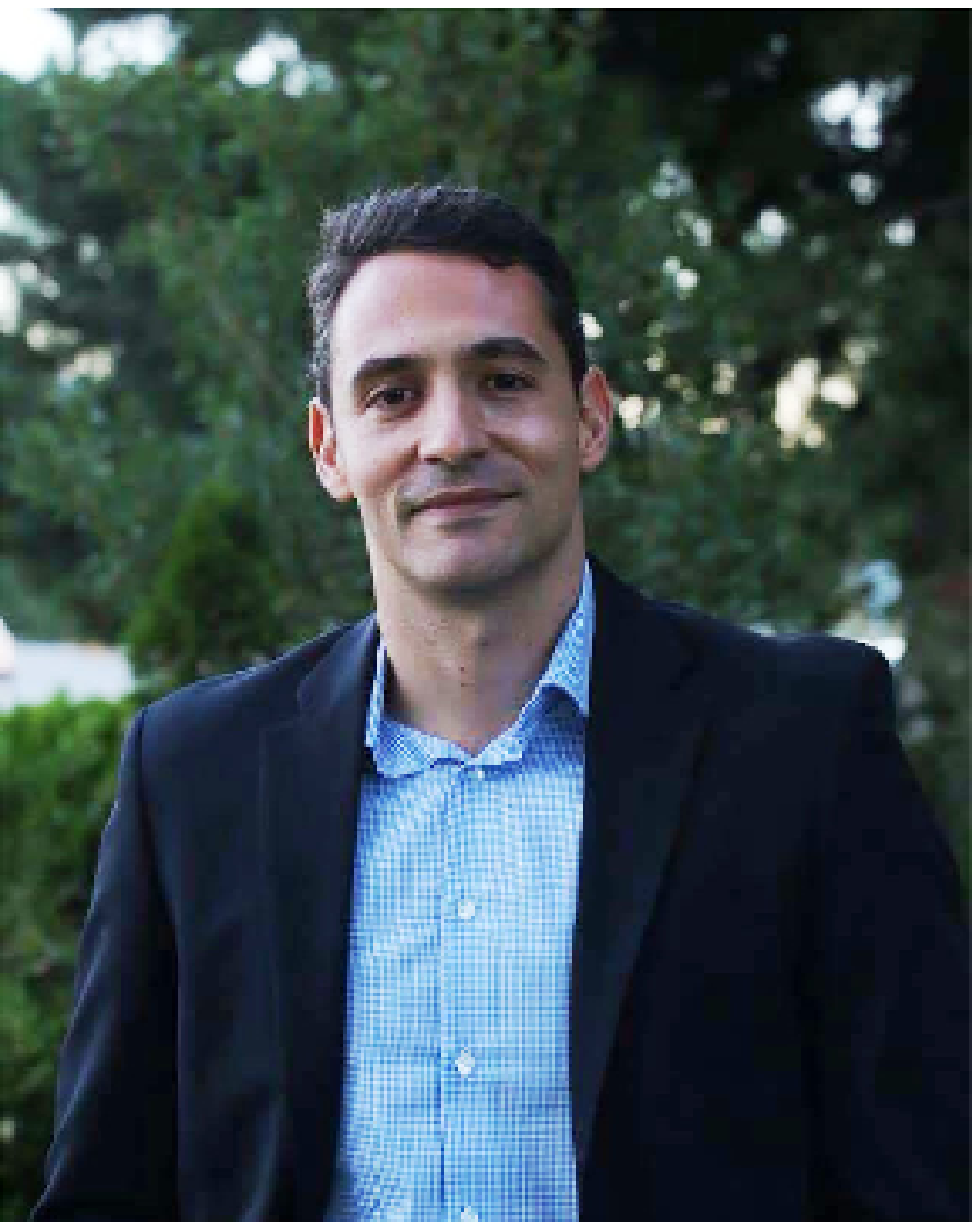} 
obtained his Ph.D. in the field of aerial robotics control and collaboration from the University of Patras, Greece in 2011. His Ph.D. research was supported by the Greek National-European Commission Excellence scholarship. Being awarded a Swiss Government fellowship he moved to ETH Zurich. From 2011 to June 2015 he held the position of Senior Researcher at the Autonomous Systems Lab, ETH Zurich. Currently, Dr. Alexis is an Assistant Professor at the University of Nevada, Reno and director of the Autonomous Robots Lab. His research interests lie in the fields of robotics and autonomy with a particular emphasis in the topics of control and planning and extensive experience in aerial robotics including the co-development of the AtlantikSolar UAV – a solar powered small aerial robot that demonstrated 81.5h of continuous flight. He is the author of more than 70 publications and has received multiple best paper awards. 
\end{biographywithpic}
\end{document}